\documentclass{article}
\usepackage{ijcai25}

\usepackage{times}
\usepackage{soul}
\usepackage{url}
\usepackage[hidelinks]{hyperref}
\usepackage[utf8]{inputenc}
\usepackage[small]{caption}
\usepackage{graphicx}
\usepackage{amsmath}
\usepackage{amsthm}
\usepackage{booktabs}
\usepackage{algorithm}
\usepackage{adjustbox}
\usepackage{algorithmic}
\usepackage[switch]{lineno}
\usepackage{enumitem}

\title{Wisdom of the Crowds in Forecasting: Forecast Summarization for Supporting Future Event Prediction}

\author{
Anisha Saha$^1$
\and
Adam Jatowt$^2$\\
\affiliations
$^1$Max Planck Institute for Informatics, Saarland Informatics Campus\\
$^2$University of Innsbruck\\
\emails
ansaha@mpi-inf.mpg.de,
adam.jatowt@uibk.ac.at
}

\begin{document}

\maketitle

\begin{abstract}
Future Event Prediction (FEP) is an essential activity whose demand and application range across multiple domains. While traditional methods like simulations, predictive and time-series forecasting have demonstrated promising outcomes, their application in forecasting complex events is not entirely reliable due to the inability of numerical data to accurately capture the semantic information related to events. One forecasting way is to gather and aggregate collective opinions on the future to make predictions as cumulative perspectives carry the potential to help estimating the likelihood of upcoming events. In this work, we organize the existing research and frameworks that aim to support future event prediction based on crowd wisdom through aggregating individual forecasts. We discuss the challenges involved, available datasets, as well as the scope of improvement and future research directions for this task. We also introduce a novel data model to represent individual forecast statements.
\end{abstract}

\section{Introduction}

The ability to correctly predict the future is of utmost importance to foresee and plan for likely outcomes. Not only does it allow people to take necessary measures, but also concerned authorities can formulate required policies and make informed decisions. 
Examples include forecasting the weather, stock-market trends, geo-political unrests, resource crises or pandemic outbreaks like COVID-19.
However, understandably, accurately forecasting the future is an extremely challenging task owing to its inherent uncertainty. While the reasoning capability and the acquired experience give humans an edge over machines when it comes to forecasting likely events, the vast amount of potentially relevant information requires computational approaches. With the recent advancement in artificial intelligence and large language models in particular, automated forecasting systems with reasoning abilities matching those of humans or beyond are not unimaginable.

There are many ways in which future forecasting could be realized.  Often, approaches essentially rely on the history of a phenomenon such as past records of stock prices, previous weather conditions over time, etc.
For example, time series-based prediction is commonly done in weather forecasting, stock markets trend analysis like options trading and in pricing. Scholars also use simulation techniques like Monte Carlo \cite{mooney1997monte} to model a sequence of events occurring in time. 
Starting from ancient times when predictions were based on trajectories of celestial bodies or astrology, the approaches to anticipate occurrence of events have slowly progressed to more refined methods like scientific modeling and advanced computational approaches. 
There is even a field of studies called 
Futurology or Future Science that aims to study past and present trends for forecasting future scenarios. 

In real life, people often resort to the opinions of others for future forecasting, such as friends, relatives or professionals to leverage collective wisdom and compare diverse perspectives. Collective judgment and verdict given by a group of people is often more reliable and accurate compared to that of an individual. ``Wisdom of the Crowd" in the forecasting context refers to the opinions and expectations that people share about a certain future event or state.
On the web, one can find abundant information related to future such as forecasts, opinions, discussions related to the future, etc. With the advent of social media, this data is more easily accessible than ever, amplifying the reach of public opinion concerning the future. Due to the relative abundance of future-related data available online, researchers started investigating approaches to aggregate individual predictions for formulating reliable forecasts. The underlying intuition is that the more often a similar event is predicted by different users, the more likely this event is to happen.

Our survey outlines the computational approaches aiming to harness the wisdom of the crowd for future prediction. We overview the concept of \emph{Future Event Prediction based on Crowd Wisdom} (\emph{FEP-CW}), which we also call \emph{Forecast Aggregation} or \emph{Forecast Summarization}, the types and sources of data collected for this specific task, the various approaches undertaken by researchers to extract temporal information from text, different methods of aggregating future-related information and finally the techniques applied to formulate and visualize predictions.

In total, 36 relevant papers were selected for our study through a keyword search in online publication databases followed by a careful analysis. We retained only those works that perform future forecasting by aggregating multiple future-related statements expressed in text. This means that researches that study patterns of historical events (e.g., frequencies, chronology, causality, etc.) to extrapolate them to the future were not included in our survey, same as works which utilize time series for making the predictions.  Our main criterion was that the relevant researches must utilize future-referring statements in text for producing the forecasts.

\begin{figure}
    \includegraphics[width=0.45\textwidth]{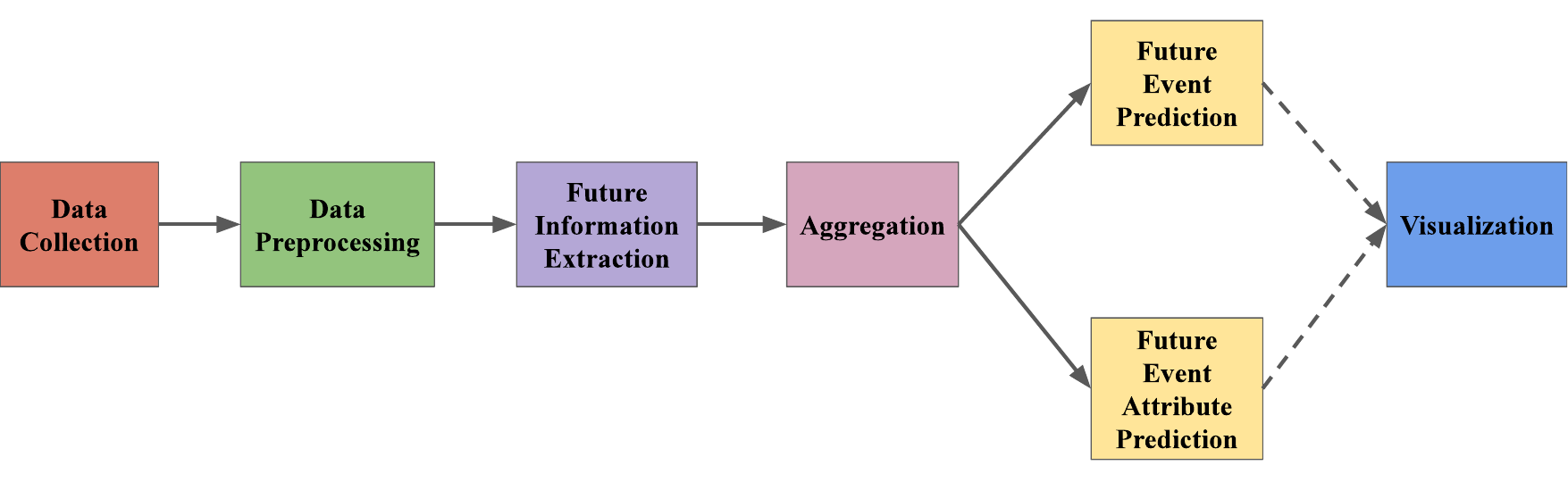}
    \caption{General Flow of Predicting Future Event} \label{flow}
\end{figure}

To our knowledge, this is the first survey on FEP-CW, despite the importance and the inherent complexity of this task, which surpasses the complexity of traditional multi-document summarization. Our focus is to provide a comprehensive deep-dive into the underlying concepts and techniques developed to target the problem of FEP-CW. Figure \ref{flow} briefly outlines the basic steps involved in FEP-CW, most of which will be discussed in detail in different sections of the paper.
Additionally, we also provide in Sec. 5 a novel data model of future-related statements useful to single out core elements that matter for the FEP-CW task. We finally list in Sec. 6 multiple different avenues for further research.

\begin{table*}
    \centering
    \adjustbox{max width=\textwidth}{%
    \begin{tabular}{lrrrrr}
        \toprule
        Dataset & Types  & Language(s) & Structured & Source & Count \\
        \midrule
        \textbf{\cite{jatowt2011extracting}} & News  & English &  No & Google News Archive & 3.6M \\
        \textbf{\cite{dias2011future}} & Web Contents & English & Yes & Web & 508 \\
        \textbf{\cite{jatowt2013multi}} & News/Web Contents & Multilingual & Yes & Web & 1,436K \\
        \textbf{\cite{hurriyetouglu2013estimating}} & Tweets & Dutch & Yes &  Twitter & 94,285 \\
        \textbf{\cite{hu2017happens}} & News & Chinese & Yes & Sina News & 155,358 \\
        \textbf{\cite{goyal2019embedding}} & News  & English &  Yes & Gigaword & 5,000 \\
        FORECASTQA \textbf{\cite{jin2020forecastqa}} &  News & English & Yes  &  LexisNexis & 10,392\\
        Autocast \textbf{\cite{zou2022forecasting}} & News Articles & English & Yes & Public forecasting tournaments &  6,707 \\
        IntervalQA \textbf{\cite{zou2022forecasting}} & News Articles & English & Yes & Various NLP datasets & 30k \\
        ExpTime \textbf{\cite{yuan2023back}} & News & English & Yes&Compilation of event forecasting datasets &26K \\
        \textbf{\cite{regev2024future}} & News& English
                &     Yes     &Longbets, Horizons, ChatGPT, New York Times &   6,800     \\
        \textbf{\cite{mutschlechner2025analyzing}} & News & English & Yes & Metaculus, Google News & 614 \\
        \bottomrule
    \end{tabular}}
    \caption{The datasets used in crowdsource-based future event prediction research.}
    \label{table:1}
\end{table*}

\section{Data and Sources} \label{data}
 In this section we discuss used document genres and datasets created for FEP-CW  task. We summarize the relevant datasets in Table \ref{table:1}. Most of FEP-CW methods leverage news articles, tweets and web pages. Some approaches make use of dedicated websites that collect predictions from diverse sources or invite users to express their opinions or cast votes. Commonly used forecasting websites like \textit{Sigma Scan}\footnote{http://www.sigmascan.org/}, \textit{Metaculus}\footnote{https://www.metaculus.com/} and \textit{Future Timeline}\footnote{http://seikatsusoken.jp/futuretimeline/} contain future information and predictions collected from news articles, research papers, government reports or through community activity.


The majority of works that perform FEP-CW opt however for news articles as their data source, likely due to their availability and up-to-date information contained about events happening around the world. An investigation by \cite{kanhabua2011ranking} revealed that nearly one-third of sentences in news articles consist of some kind of reference to the future. News articles have been then commonly used for creating datasets for FEP-CW. Additionally, some researchers \cite{nakajima2014investigation,radinsky2013mining,nakajima2020future} advocate directly retrieving Web content instead of relying on any prepared datasets or collections.

We discuss below a few of the commonly used dataset formats. In the FORECASTQA dataset \cite{jin2020forecastqa}, English news articles are converted to ${<Question, Answer, Timestamp>}$ triples for answering binary and multiple-choice forecasting questions. However, FORECASTQA suffers from ambiguity and the lack of context, owing to it being crowdsourced. The unstructured nature of this dataset requires models to additionally link relevant events to answer forecasting questions. Addressing these challenges, \cite{zou2022forecasting} released the Autocast (forecasting questions of type True/False, Multiple-Choice or Numerical) and IntervalQA (questions having numerical answers only) datasets. Recently, \cite{mutschlechner2025analyzing} curated a binary forecasting questions dataset to evaluate performance of LLMs. \cite{goyal2019embedding} built a dataset containing temporally ordered event pairs - $<sentence1> <sentence2> <event1> <event2> <temporal relation>$. The TimeLlaMA \cite{yuan2023back} model for explainable event prediction is based on a multi-source instruction tuning dataset, ExpTime with documents, forecasting questions, predictions and explanations for these predictions. \cite{regev2024future} built a dataset containing manually labeled future and non-future related sentences. Besides English news and tweets, some works \cite{hu2017happens,jatowt2015mapping,hurriyetouglu2013estimating,tops2013predicting,jatowt2013multi} harvested multilingual data. 



Curating datasets for FEP-CW involves often extensive preprocessing, e.g., with commonly available toolkits like OpenNLP for sentence splitting, tokenization, part-of-speech tagging, and shallow parsing, TARSQI Toolkit for annotating documents with TimeML, SuperSense for named-entity recognition, etc. Documents retrieved from the Web may need to be also filtered out for removing duplicates. 

An important problem with FEP-CW datasets is that they can quickly become obsolete. Not only the patterns of events might change over time causing models trained on outdated datasets exhibit lower accuracy, but there is a risk of future information leakage. Expecially, when it comes to Large Language Models (LLMs), their training cutoff dates need to be carefully compared against the dates of events to be forecasted to minimize the risk of models already knowing the results of these events \cite{mutschlechner2025analyzing}.
Otherwise, the evaluation would not be fair.  
In fact, accumulating future-related content could be considered as an integral part of the task as in some working systems \cite{regev2024future}. This however makes scientific reproducibility more challenging.


\section{Extracting future-related information} \label{temporal}
Usually, documents are not purely about the future, but rather future-related content is mixed with the one referring to the past or present. For example, in news articles, statements about planned events or forthcoming course of actions appear together with the content reporting ongoing or recent events.
While the forecast summarization can, in principle, be done on entire documents (e.g., by prompting LLMs to focus on future information only), often a common approach is to first extract future-related statements to be later used in aggregation and summarization. 

A simple way to collect relevant data is by using standard search techniques by issuing pre-specified sets of keywords.
In general, conventional search engines have been used for time-oriented ranking since already some time ago \cite{DBLP:conf/cikm/AlonsoGB09,jatowt2005temporal,campos2017identifying,dias2011future}, and this also applies to ranking results so that future-related ones get returned at the top of search results list. \cite{baeza2005searching} was one of the first to introduce the concept of a search engine for extracting future temporal expressions from news articles. They investigated almost half a million future-referring sentences from Google News and inferred that the closer the event forecasted in the prediction is to the prediction's creation time, the higher is the confidence and accuracy of the prediction.





\subsection{Extraction based on Time Expressions} 

When the publication time of a document is known (as in news articles), it is usually easy to find future-related information if it is accompanied by explicit dates. Sentences containing embedded temporal expressions or references that point to the time later from the document publication date are then identified as future-referring instances wrt. the document. We call these \emph{per-document future-related statements}.
Whether they are actually future-related at a given inference time needs to be determined by comparing them to the time point of ``now". Per-document future-related statements that are no longer referring to the actual future  can however be still useful. For example, they can be used for training extraction models. 

 A temporal expression can be defined as a sequence of words that denote a point in time, a time interval or the frequency of occurrence of an event.  
 Temporal expressions after being detected in text may need to be normalized. There can be explicit, relative, or implicit temporal expressions. Explicit or absolute temporal expressions directly indicate a time point or an interval, for example, \textit{``1st April 2024"}. Relative temporal expressions refer indirectly to a point of time, by means of expressions like \textit{``two weeks later"}, \textit{"a month ago"}, etc. Implicit temporal expressions refer to events like, "The Asian Games" that signify a time event. 
 
 We emphasize that due to the multiple senses of words and overall language ambiguity, extracting and normalizing time expressions is not always trivial and often needs to be done using specialized temporal taggers \cite{mani2000robust,strotgen2010extraction} if high accuracy is important. \cite{mani2005language} created the TimeML framework through annotations of temporal events, marking the commencement of frequent use of text data for understanding the underlying relations between time and events. 
 Built on top of the TimeML framework are commonly used temporal taggers like GUTime \cite{mani2000robust}, HeidelTime \cite{strotgen2010extraction}, Stanford CoreNLP tagger and SUTime. However, these taggers possess some limitations. \cite{regev2024future}, for example, reported that SUTime failed to perform in ambiguous cases. For instance, it maps \textit{"On Tuesday"} to the upcoming Tuesday, rather than the past Tuesday referred to in the sentence \textit{"On Tuesday, he revealed his decision, which will become official soon."}.  

%

Simple patterns or regular expressions involving explicit temporal expressions can be used as an alternative to temporal taggers, especially when one wants to extract large number of future-related statements without high computational cost. \cite{jatowt2010analyzing} in their exploratory analysis collected over a million time-referenced future-related statements from the Web in English, Japanese and Polish by searching using structured queries with temporal expressions defined as \textit{$temp_{modifier}+(the)year(s)+yyyy$} (where, \textit{$temp_{modifier}$} is a preposition) for retrieval by years, which translates to expressions like \textit{“in year 2027”, “in the year 2035”, “by the year 2049”} and so on. Similarly, for retrieval by months, the expressions was based on \textit{“$month\_{reference}+yyyy$”} where \textit{$month\_{reference}$} is a full or shortened form of the name of a month (like, February or Feb). 

In a similar way, ChronoSeeker \cite{kawai2010chronoseeker} retrieves both past and future time-referenced events that are relevant to the user's query and filters noisy events using support vector machines. The system has five modules: search (a search API returns year expressions $y$ relevant to a user query $q$), extraction (extracts candidate sentences containing $y$), filtering (a machine-learning technique which filters these candidates), clustering (groups sentences related to the same event) and visualizations (timelines of chronological events).
 To analyze the impact of temporal features on the classification of different types of Web snippets, \cite{dias2011future} employed a pattern-matching methodology as proposed in \cite{campos2011temporal} which focuses on years. They investigated whether a unigram model combined with temporal features can classify future-related web snippets into one of the genres (informative, scheduled or rumours) for five different classifiers, only to conclude that depending on the learning algorithm temporal features might improve future-related text classification to a great extent. 

While the time expressions-based extraction is characterized by a relatively high precision, the main problem with this approach is low recall. This is because many future-related statements lacking explicit temporal pointers are omitted which causes bias in the data collection towards more established or concrete forecasts and plans.

\subsection{Extraction based on Future Indicating Keywords}
Forecasts with specific time markers may be for relatively nearer and more certain events in the future. The absence of an explicit mention of any date or even a vague time pointer makes it more difficult to retrieve future mentions. The implicit future-related statements need then to be found through other means.  

 One way is to look for the occurrence of certain markers of future-relatedness like \textit{``will"} or \textit{``future"}, although many other phrases like \textit{``...plans to"} or \textit{``...hope to"} might serve as well. 
Searching with the keywords like exemplified above might yield some results but carries a high risk of returning more false-positive cases. For example, a sentence like \textit{"He has a strong will to pursue higher studies"} might be misleading. Additionally, the list of keywords needs to be extensive for this approach as there can be many different ways one can refer to the future, including the use of tenses other than the future tense.

\cite{kanazawa2011improving} proposes a simple technique to determine future orientation markers. 
They employ statistical test to detect characteristic terms that frequently appear in a large collection of future-related statements while being infrequent in the collection containing content related to the past. The two content collections are formed by relying on temporal expressions as discussed in the previous section.
Having determined characteristic terms that often appear in future-related sentences, they proceed to extracting implicit future-related statements.
The sentences are ranked based on the average weights of the characteristic terms in their content. A sentence that contains many highly-scored future characteristic terms and few highly-scored past characteristic terms is more likely to refer to a future event. 

A better way than the statistical analysis is to use trained neural network classifiers like DistilRoBERTa as proposed in \cite{regev2024future}. However, the problem lies in the fact that this approach is not very efficient when searching for large amounts of future-related content, but rather as a post-processing tool to filter out non-future-related content.

\subsection{Extraction based on Morphological Patterns}
The complexity of natural language might mislead a model to perceive future information as a past reference. Consider the sentence, \textit{"Carol has decided to open a new bakery"}. Despite the use of past tense in \textit{"decided"}, the sentence indicates an upcoming event. To parse complex future expressions as this example, sentences can be represented in a morphosemantic structure \cite{levin2017morphology} which is a combination of semantic role labeling and morphological information.

\cite{nakajima2014investigation} introduce a sentence extraction system \textit{SPEC} based on semantic role annotations for extracting frequent sentence patterns from a corpus. The approach is based on the intuition that there are multiple possible future expressions while some words or phrases are used as future expressions only in a specific context. Hence, there must exist characteristic patterns (grammatic or semantic) appearing in future-related sentences that distinguish them from non-future referring sentences. 
\cite{10.5555/2832747.2832885} investigates patterns in multiple news corpora, that are semantically and grammatically consistent, yet lexically different. They propose a method for automatically extracting such frequent patterns and establish its effectiveness on the downstream task of identifying future-related sentences. \cite{nakajima2020future} then use frequent combinations of patterns for classifying future-referring sentences in Japanese by using MeCab analyzer to extract morphological information for Japanese words. \cite{ni2015computational} and \cite{al2018automatic} apply similar approaches to analyze future-referring expressions in English and Arabic, respectively. 

\subsection{Extraction using Language Models}
LLMs have excelled recently in natural language understanding (contextual, semantic, etc.) and generation. 
As trained on huge text corpora, the models naturally   
observe a variety of ways in which opinions on future are expressed throughout diverse document genres (news articles, blogs, scientific papers, etc.). 
\cite{regev2024future} propose using encoder only language models, such as the DistilRoBERTa model \cite{liu2019roberta}, as classifiers to distinguish future and non-future referring sentences. The authors use 6,800 manually labeled sentences for training to extract positive sentences with a strict confidence threshold of 0.9. 
 In case of larger language models, special prompts need to be designed to elucidate information about the future. A study by \cite{yuan2023back} shows that directly prompting ChatGPT \cite{ouyang2022training} to identify future-related content leads to suboptimal accuracy. 




\subsection{Post-Filtering Past Future}
A potential subsequent step in the data collection/extraction process is to validate the predictions by ensuring they have not already occurred. The predictions that are no longer valid should be removed from the collection. This filtering step, often skipped, could play a crucial role in assuring the quality of generated forecast summaries.
\cite{kanazawa2011improving} proposed a simple two-stage method for estimating validity of the predictions by searching for relevant events that have already actually occurred (or their forecasted horizon have already passed) and calculating cosine similarity between the forecasts and the gathered events.
 
 Note that the above concept of the "already passed future" is different from the case when an earlier prediction gets invalidated or contradicted by a later one (e.g., a company officially canceling its previous plans). Those resolutions may need to be accounted for during the aggregation stage.

\section{Aggregating Forecasts} \label{event}
Event prediction methodologies vary due to the underlying heterogeneity of the prediction scope, which might range from the prediction of only a single attribute of an event like date, time, location, actors, etc. or their combination, to the prediction of an event itself or even a sequence of events. 
Additionally, the actual form of output might vary from a phrase, a sentence, a collection of sentences, binary answer (True or False) such as for questions whether an event will happen, a timeline, etc.  
Some works even formulate the problem of event forecasting as a problem of link prediction in knowledge graphs or temporal knowledge graphs \cite{ma2023context,lee2023temporal,ma2023structured,deng2020dynamic}.



\subsection{Frequency-based Future Analysis}
Certain approaches do not predict individual events but rather focus on presenting the overall statistics of future-related content such as frequency, average sentiment, common words, etc. \cite{jatowt2013multi} conduct a large-scale exploratory analysis of future-related information on the web studying their temporal horizons, common keywords, and sentiment degrees in three different languages. \cite{jatowt2015mapping} analyze a large collection of tweets to quantify temporal attention and related temporal characteristics expressed by users by representing time mentions as a probability distribution over a time duration.

\subsection{Forecasting through Clustering}
Due to the lack of dedicated and annotated datasets, many approaches resort to unsupervised ways of forecast aggregation such as clustering.
The clusters could be formed based on semantic similarity of input forecasts, their temporal similarity, which typically denotes the agreement between the expected dates of predicted events, or a combination of both. \cite{jatowt2009supporting} clusters predictions related to input query using k-means containing future temporal expressions based on the mixture of content and time similarities and then selects the best clusters to summarize probable future events. The later post-clustering step is needed to discard semantically incoherent or temporally elongated clusters to eliminate noise. 

\cite{jatowt2011extracting} undertook a mixture-model-based clustering approach to estimate the probability of a future event by grouping related predictions. When a user queries a topic, a cluster is created based on the related events' textual and the forecast horizons are used to form a probability distribution over time. The authors use logarithmic timeline which collapses time further ahead mimicking human style to think about the future in which distant events are represented by coarser time granularity than near events. A more recent approach by \cite{regev2024future} uses BERTopic library to extract topics based on similarity of forecast embeddings.

\subsection{Forecasting through Ranking}

Ranking future predictions can be considered as another way of forecast aggregation and selection.
\cite{kanhabua2011ranking} proposes an approach for automatically generating queries from a news story read by a user to retrieve predictions related to that news article. 
The predictions are then ranked by their relevance (future information) and are returned to the user. The ranking algorithm is built on a support vector machine framework, based on features like term, entity, topic and temporal similarity which captures the correspondence between the generated query and predictions. This work is actually an example of an interesting application of FEP-CW as the trigger in this case is a document being read by a user unlike other approaches that use entities as their input. 

\subsection{Forecasting using LLMs}

With LLMs becoming very usefil, it is natural to study the effectiveness of LLMs in forecasting, effectively, using them as a kind of aggregation and reasoning tool. With this intention, \cite{schoenegger2023large} evaluated GPT-4's \cite{achiam2023gpt} capability of future prediction. The authors concluded that GPT-4’s forecasts are significantly less accurate compared to the median human-crowd forecasts.  They also hypothesized that a potential reason for the underperformance is that in making real-world forecasts the actual answers are unknown, unlike in other benchmark tasks where the strong performance of an LLM could be (at least partially) due to the answers being memorized from the training data. 

\cite{mutschlechner2025analyzing} found out that supplementing background context in the form of related news articles or future-related question's resolution criteria combined with an appropriate prompt structure can improve the performance of even older LLMs like GPT-3.5 turbo.
\cite{pernegger2024evaluating} investigated the performance of BERT \cite{devlin2019bertpretrainingdeepbidirectional}, RoBERTa \cite{liu2019robertarobustlyoptimizedbert} and GLM \cite{du2022glmgenerallanguagemodel} and concluded that zero-shot learning did not improve the language models' forecasting abilities and that retrieval-augmented generation (RAG) offers more chance for substantial improvement. Recently, \cite{zhang2024largelanguagemodelsevent} formulated the problem of multi-event forecasting as predicting occurrences of multiple unique relations in a temporal knowledge graph (TKG). They design a prompt template such that it does not exceed the maximum permitted token limit for open-source LLMs while including as much historical information as possible. Further, they utilize the pre-trained LLM, RoBERTa-large to encode this prompt and employ a self-attention-based prediction head to handle the output embeddings and predict the relation indicating the future event occurrence. \cite{nako2025navigatingtomorrowreliablyassessing} investigated the effectiveness of LLMs for likelihood questions by asking if an event is likely to happen instead of if it will happen. For better accuracy, the authors also created fake events probing how these will be predicted by LLMs.
\vskip \baselineskip

\begin{figure}[!tbp]
  \centering
    \includegraphics[width=0.4\textwidth]{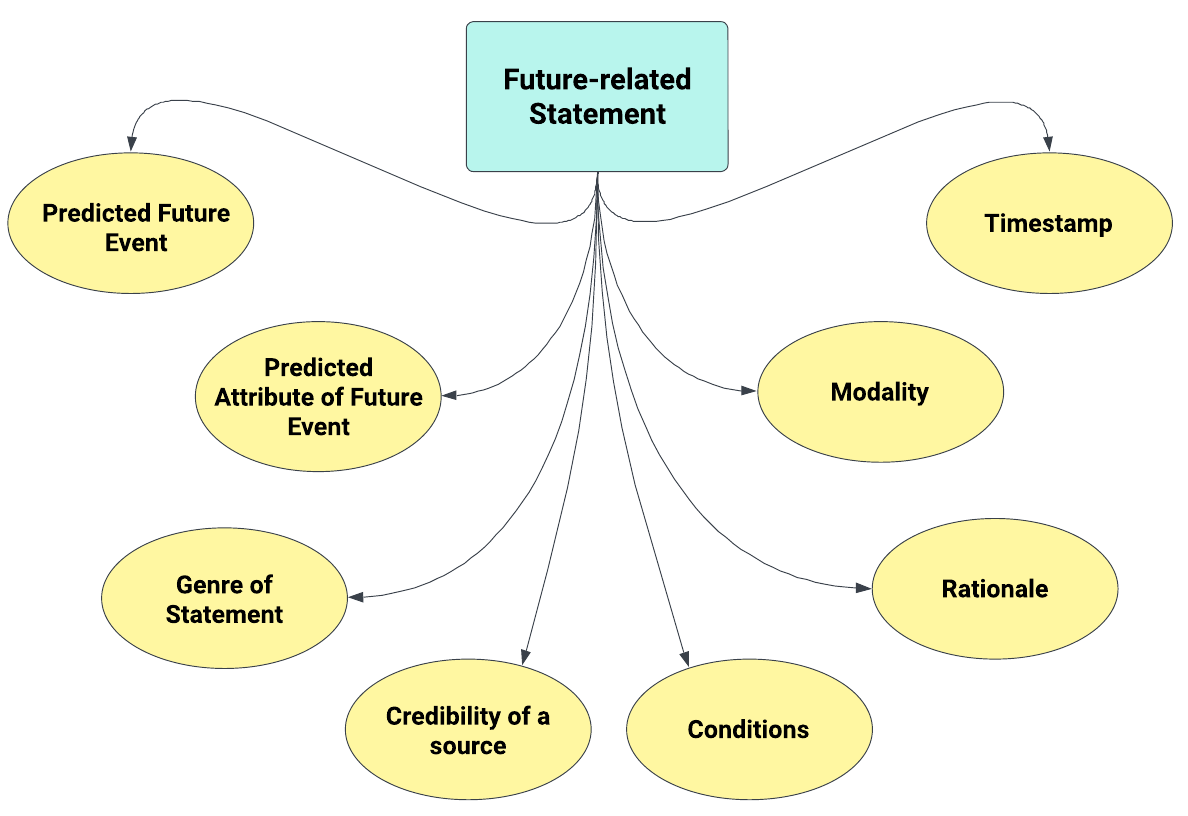}
    \caption{Atomic Components of Individual Future Prediction that can be harnessed for FEP-CW.} \label{comp}

\end{figure}

\begin{table*}
\centering
\small
\vskip\baselineskip
\begin{tabular}{|p{5cm}|p{7cm}|}
\hline
\multicolumn{2}{|c|}{\parbox{12cm}{\vspace{0.5\baselineskip}\textbf{FRS}: (July 31, 2024) Japanese National Bank will try to increase the value of Yen after October since the country needs to import many goods and their current price is too high. This however would depend on the actions of FED and the changes in the interest rate in the USA.\vspace{0.7\baselineskip}}} \\
\hline
\textbf{Component} & \textbf{Example}\\
\hline
Predicted Future Event & Increase the value of Yen.\\
\hline
Predicted Attribute of Future Event &  Date: After October 2024.\\
\hline
Genre of Statement & News article\\
\hline
Credibility of a Source & High\\
\hline
Conditions & Actions of FED and the changes in the interest rate in the USA.\\
\hline
Rationale & The country needs to import many goods and their current price is too high.\\
\hline
Modality & "will try to" (moderately plausible)\\
\hline
Timestamp & July 31, 2024\\
\hline
\end{tabular}
\caption{Example of the proposed data model components}
\label{table:2}
\end{table*}

\vspace{-1em}
\section{Data Model of Future-related Statements}
We outline in this section our proposal of a data model that decomposes a future-related statement into several distinct components. These components can be useful for a more effective aggregation/summarization process using refined methods. Figure \ref{comp} shows the breakdown of a future-related statement into components that can play key roles in FEP-CW, described as below. Table \ref{table:2} also highlights these components through an example. 
\begin{itemize}
    \item \textbf{Predicted Future Event} refers to the description of an event expected to occur by the statement creator.
    \item \textbf{Predicted Attribute of Future Event} are any associated information about the future event that the statement creator forecasts in her future-related statement (e.g., actors, location, result, consequences). Of particular importance is here the attribute that specifies the \textbf{date/time of the forecasted event} or a forecast horizon. 
    \item \textbf{Genre of Statement} defines the type of a document in which the future-related statement appears setting up the general context for understanding the statement. News, web and social media are a few of the common sources.
    \item \textbf{Credibility of a Source} determines the quality and reliability of the statement creator (an expert or average user, an institution, etc.) which helps to assess the quality and veracity of the future prediction. Note that the credibility is strongly affected by the relation of a source to the forecasted events (e.g., a company expressing its own future plans vs. an investor or a customer making the prediction about this company).
    \item \textbf{Conditions} are usually specific circumstances or constraints that need to occur or be satisfied for an event to happen. These constitute other future events expressed as embedded future-related statements, potentially, with their own components such as respective dates, modality, or even their own conditions\footnote{This essentially leads to a sequence or hierarchy of conditions.}. Note that there could be multiple conditions in a single future-related statement.
    \item \textbf{Rationale} defines the logic behind why a given future event is expected to happen, usually in the form of an explanation supporting why the predicted event will happen. 
    Sometimes however rationale may only contain expressions of modality or subjective confidence without backing them by any logic or reasoning. 
    \item \textbf{Modality} indicates the certainty level of a future event made by the statement maker and can be expressed using words like \textit{"will", "likely", "might",} etc. Any future-related information has some level of uncertainty and the modality determines how the statement creator assesses the likelihood of the forecast to happen. Modality can be also in the form of a  specific expression of confidence (e.g., 60\%).
    \item \textbf{Timestamp} denotes the date or time when the particular future-related statement has been made. A general assumption is that the older the statement, the less probable is that it will be still correct (or still valid). 
    The more recent predictions can be thus preferred in case of \emph{forecast conflicts}, or they may have  higher weights assigned for the aggregation step. 
\end{itemize}


The above-described model may form the base for training dedicated classifier(s) that would allow extracting individual components. These can be then utilized in various ways when aggregating individual forecasts (e.g., weighting them by timestamps, credibility, expressed modalities, a presence or argumentativeness of associated rationale or likelihood of necessary conditions to be fulfilled). Note that not all of the above-listed elements may be stated in a particular future-related statement or may be known. For example, when expressing predictions, one may not mention (or even know) the date when the event will occur or there might be no particular prerequisites given in the form of conditions. However, when available, some of these elements can be quite useful for creating the final forecast summaries such as the rationale or credibility of a statement producer which can help determine the accuracy of predictions.



\section{Future Directions} \label{challenges}
 In this section, we identify potential future research directions. 

\begin{itemize}
    \item \textbf{Approaches using Future-related Data Model}: Currently, few approaches make use of the individual elements explicated in the data model in Sec. 5. Most just treat predictions as atomic units or consider only more obvious elements such as timestamps. The model can be used to more accurately understand predictions by individual actors and can lead to their better combination. For example, predictions could be grouped not only by the similarity of the forecasted events but also by the similarity of their conditions or rationale to more accurately reflect their support within the community. Another way is to assign weights to individual forecasts based on their source credibility, genres, timestamps, complexity of conditions, logic of rationale, etc.
    
    \item \textbf{Multilingual datasets}: Creating datasets across languages from different regions will enable incorporating local context and lead to more nuanced understanding of model capabilities. 
    \item \textbf{Domain-specific datasets}: Collecting domain-specific data will add knowledge from particular areas like agriculture, business, and health and help in generating specialized real-time forecasts.
    \item  \textbf{Resolving forecast conflicts}: Since future is inherently uncertain one can expect many forecast conflicts which can be automatically detected and resolved, e.g., with the help of assessing prediction validity and timeliness.
    \item \textbf{Evaluation and metrics}: There is a need to formulate evaluation benchmarks, both generalized and domain-specific and formulate context-aware metrics that will help in better evaluation. Currently, evaluation is either missing, or based on a manual verification if the predicted events happened.
    \item \textbf{LLMs for forecasting}: GPT-4 has been found to underperform in real-world-forecasting, event if supplied with background information about an event \cite{yuan2023back}. In the future, LLMs' forecasting capabilities can be improved by accessing real-time information (RAG) from the web for mitigating knowledge cutoffs, integrating human-in-the-loop, adapting chain-of-thought prompting, or investigating better prompting techniques and the role of supporting, contextual information  \cite{mutschlechner2025analyzing}.
    \item \textbf{Integrating wisdom of the crowd approach with other forecasting methods}: Combining public opinion analysis and other forecasting approaches such as time series based forecasting, or causal analysis is a promising area.
    \item \textbf{Bias removal}: Future forecasting faces the risk of bias. For example, statements originating from a single source may be skewed towards certain event outcome or certain interpretation of the future. It is important to maintain carefully the diversity of used sources and consider application of bias detection techniques 
\cite{farber2020multidimensional,10140917,gallegos2024bias}.
    \item \textbf{Multimodality}: Other modalities in addition to text can be incorporated in the prediction task (e.g., images, time series, videos).
    \item \textbf{Creating future timelines}: Timeline summarization techniques \cite{yu2021multi,chen2019learning} 
    can be adapted to future forecasting outputting chronologically arranged sequences of events potentially with explained causal relations between them (e.g., connecting future events by event-conditions relations).
\end{itemize}

\section{Conclusion}\label{conclusion}
Future forecasting is inherently challenging but, at the same time, it is a very common activity that humans do on a daily basis both in their private and professional lives. Forecasting is also of great importance for any domains and business. It is then important and natural to study how to harness computational approaches for enhancing future forecasting, especially, considering the impressive advancements in NLP methods we recently witness.

We are the first to survey previous research works that follow a particular way to generate predictions about the future - by attempting to detect, aggregate and summarize individual forecasts issued by multiple users. The application of the Wisdom-of-the-Crowd concept for future forecasting is appealing given the abundance of opinions shared online including the expressed opinions on likely future. 

In our survey, we overview the techniques, challenges, data sources and promising avenues of research in the task of future event forecasting using the wisdom of the crowd. Our work details the data sources and types, methodologies, and future directions. We also introduce a novel data model that decomposes future-related statements into separate elements that carry potential for effective aggregation and for subsequent generation of effective predictions. In the future, we plan to extend this data model  and conduct experiments to determine the extraction performance. 

\textbf{Challenges}. 
Our survey could be useful for scholars as the first step when starting to approach this challenging task or practitioners who want to develop intelligent systems for providing real-time applications.
While problem-solving techniques in NLP have evolved over time, a comparatively under-explored domain as this faces numerous challenges. A key problem is with the evaluation. While a natural way is to verify the predictions based on real outcomes, this means that either one needs to wait to check if the forecasts' results were indeed correct, or to use past forecasts with already known event outcomes. The latter approach carries the risk of data contamination especially when using LLMs.
This means that datasets quickly become obsolete.
The task suffers also from the lack of structured datasets not only in English but across other languages.

\bibliographystyle{named}
\bibliography{v2}

\end{document}